# PLM-GNN: A Webpage Classification Method based on Joint Pre-trained Language Model and Graph Neural Network

Qiwei Lang[1], Jingbo Zhou[1], Haoyi Wang[1], Shiqi Lyu[1], Rui Zhang[2]*

[1] College of Software, Jilin University, Changchun, China
[2] College of Computer Science and Technology, Jilin University, Changchun, China
Email: {langqw5520,zhoujb5520,wanghy5520,lvsq5520}@mails.jlu.edu.cn,
rui@jlu.edu.cn

**Abstract.** The number of web pages is growing at an exponential rate, accumulating massive amounts of data on the web. It is one of the key processes to classify webpages in web information mining. Some classical methods are based on manually building features of web pages and training classifiers based on machine learning or deep learning. However, building features manually requires specific domain knowledge and usually takes a long time to validate the validity of features. Considering webpages generated by the combination of text and HTML Document Object Model(DOM) trees, we propose a representation and classification method based on a pre-trained language model and graph neural network, named PLM-GNN. It is based on the joint encoding of text and HTML DOM trees in the web pages. It performs well on the KI-04 and SWDE datasets and on practical dataset AHS for the project of scholar's homepage crawling.

## 1. Introduction

Web page classification is one of the classic tasks in the process of web information mining. There are two major categories of solutions, manual web classification, and automatic web classification[1]. Manual web classification is the task for the domain experts to classify manually based on their knowledge of the domain[2]. Automatic web classification is a supervised learning problem that requires the construction of features that can significantly distinguish the web pages, then training a classifier by some means based on the labels. Obviously, manual web classification is a tedious and time-consuming task. Although it may take some time to build and train the classifier, the latter is better than manual methods.

We propose a novel classification method, PLM-GNN, which is based on a joint pre-trained language model(PLM) and graph neural network(GNN). We consider the automatic web classification task as a two-stage process. Firstly, it constructs a representation of web pages. Secondly, using the representation to train a classifier. Some past approaches usually build features based on text or visual features and links between web pages. Text features are usually obtained using methods such as TF-IDF, Word2Vec, etc. Visual features are usually obtained through the calculation of coordinates. While links between web pages are used to construct features from other pages based on the assumption that interlinked pages are most likely to have similar features and thus migrate to the current page. However, the number of pages continues to increase, the variability of pages becomes larger. So we think that the links between pages can no longer be used as the basis for feature construction. Due to the proposed Transformer architecture and the advent of PLM, contextual information can be better learned, so we can get a better representation of the text. Our PLM-GNN is to process the text contained in web pages by PLMs. Since the structure of a web page can be reflected by the HTML DOM tree, and the structure of similar web pages is resemble, constructing the structure feature of the DOM tree should also be a key aspect. Considering that a tree is a directed acyclic graph, we introduce GNNs to learn the graph structure. We use a multi-layer perceptron(MLP) to build the final classifier to ensure the completeness and trainability of the model.

Overall, we make the following contributions:
- We introduced pre-trained language models to get a better representation of text contained in the web pages.
- We introduced graph neural networks to learn HTML DOM tree features and get a representation of the DOM tree.
- Our model constructs the features automatically, without manually constructing interactions.

## 2. Related Works

### 2.1. Text Representation

Text representation is a problem of studying how to turn a string into a vector and how well the vector can respond to the text features. The existing models are broadly classified into three categories: vector space-based models, topic-based models, and neural network-based approaches. Vector space-based methods have simple models and clear meanings, but they cannot handle synonym and near-synonym problems well. Topic models try to implement the representation of text from the perspective of probabilistic generative models, where each dimension is a topic. However, topic models suffer from problems such as long training time due to many training parameters and poor modeling of short texts. With the rise of deep learning, neural network-based representation methods have made a big splash in NLP research. Mikolov et al. proposed Word2Vec, Doc2Vec, fastText. Later, recurrent neural networks were proposed, such as RNN and LSTM. In recent years, the emergence of models based on attention mechanisms such as BERT[3], GPT[4], etc. has refreshed the baseline of NLP for various tasks.

### 2.2. Graph Neural Networks

Graphs describe pairwise relations between entities for real-world data from various domains, which are playing an increasingly important role in many applications. In recent years, GNNs have achieved tremendous success in representation learning on graphs. Most GNNs follow a message-passing mechanism to learn a node representation by propagating and transforming representations of its neighbors, which significantly helps them in capturing the complex information of graph data[5]. The most classical methods in GNNs are Graph Convolutional Networks(GCN)[6] which is based on spectral domain and Graph Attention Networks(GAT)[7] which is based on spatial domain. For GCN, the node features on the graph constitute the graph signal. For GAT, the effect of weighted convolution is achieved by computing attention scores on node pairs with the help of a self-attention mechanism.

## 3. Problem Formulation and Approach

### 3.1. Problem Formulation

In a web page classification task, we are trying to map input web pages into discrete categories. Here we focus on single-label classification tasks. Let $C = c_1, c_2, \cdots, c_M$ be a set of pre-defined categories, where $M \geq 2$. Given a set of web pages $W = w_1, w_2, \cdots, w_N$, generally speaking, we expect to find a function $f: C \times W \to C$ that can be obtained through learning to approximate the real assignment function $f_0: C \times W \to C$. The function $f$ is called a classifier or a model.

As introduced in Section 1, we regard the web page classification task as a two-stage job. Usually, we are able to get the features $X$ of a page $w \in W$ in the first stage, either by manually constructing features or by some way of learning. Let function $\varphi$ denote the mapping from a page to its features, i.e. $\varphi(w) = X$, assuming $X \in \mathbf{R}^n$. Secondly, we try to train a classifier $\gamma$ whose input is the features obtained in the first stage and output is its label, i.e., $\gamma(X) = c$, where $c \in C$.

The task in this paper is formalized as to find a representation function $\varphi$ and the best classifier $\gamma$, where

$$c = \gamma(\varphi(w)), \quad \text{for } w \in W,\ c \in C$$

to approximate the real assignment function $f_0$, where $\varphi: W \to \mathbf{R}^n$ and $\gamma: \mathbf{R}^n \to C$.

*3.2. Approach Review*

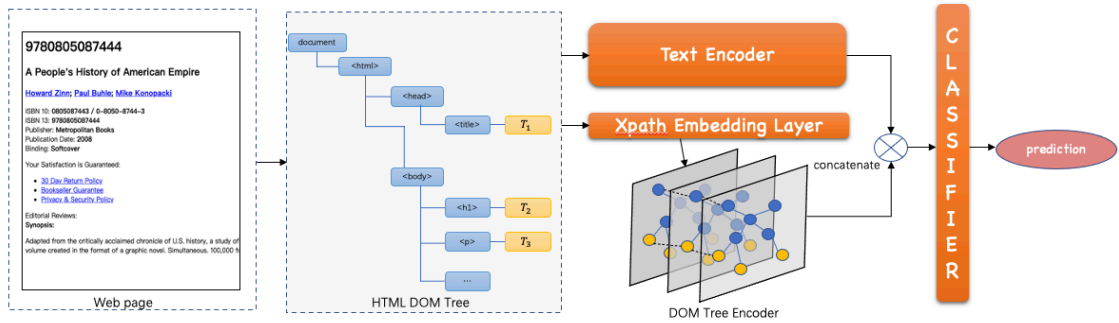

**Figure 1.** Approach Overview

Figure 1 shows the overall framework of the proposed PLM-GNN model for the web page classification task. We first parsed the web page into an HTML DOM tree. After that, we use the DOM tree to get the text information and DOM tree structure information respectively. Note that the text information is only on the leaf nodes of the DOM tree, so we get all the text information of the web page by directly traversing the whole DOM tree. We feed the text into a text encoder to get a representation of the web page text, where the text encoder is implemented by PLMs. For the DOM tree structure information, only the skeleton structure of the tree is available at the beginning. So we first construct the graph structure of the whole DOM tree. We chose the XPath as the information of each node. We train the XPath embedding layers to obtain the representation. After that, we fed the graph structure and the representation of the nodes to the graph encoder for learning. Here the graph encoder is implemented by GNNs. Since we eventually expect the representation of the whole DOM tree, we read out each graph by a global pooling strategy. After the above process, we concatenate the text representation and the DOM tree representation to get the representation for a web page. We input the vector into an MLP classifier and train it under a multi-classification task.

## 4. Encoder and Classifier
The model consists of three components: the text encoder, the DOM Tree encoder, and the MLP-based classifier.

*4.1. Text Encoder*

We first obtain all the text on a web page. Given a web page $w \in W$, we first clean the original HTML document to erase some useless tags. Then we parse the page into an HTML DOM tree $\mathcal{T} = (\mathcal{N}, \mathcal{C}, \mathcal{R})$. $\mathcal{N}$ denotes the nodes set in the DOM tree, which represents all the tags in HTML. $\mathcal{C}$ denotes the contents contained in an HTML document, especially referring to the texts $T$ in HTML. Obviously, $T$ is a subset of $\mathcal{C}$. $\mathcal{R}$ denotes the relation between nodes in $\mathcal{N}$, which usually includes parent-child and sibling relationships. We stipulate that the relationships in $\mathcal{R}$ are directionless.

After parsing the DOM tree $\mathcal{T}$, we would like to extract texts $T$ in HTML. Since the text exists on the leaf nodes of the DOM tree, we find the leaf nodes by traversing the DOM tree and then extracting all the text in the leaf nodes. Assuming leaf node $\mathcal{N}_i$ has text $T_i$, then $T$ can be obtained by concatenating all $T_i$, i.e., $T = \|_i T_i$.

We use PLMs to do the vectorization of the text. We use the tokenizer to divide the text into token sequences, i.e., $T = [w_1, w_2, \cdots, w_{L_0}]$, where $L_0$ denotes the length of the token sequences. Considering that all the text contained in a web page may be too long, however, some PLMs such as BERT have input length requirements. Let $\eta$ denote the truncation and padding operator, i.e., $\eta(T) = [w_1, w_2, \cdots, w_L]$, where $L$ denotes the length required by the PLM.

Finally, we feed the input sequence $\eta(T)$ into a PLM, then we got the text representation $\mathbf{x}_T$ through

$$\mathbf{x}_T = \text{PLM}(\eta(T))$$

where $\mathbf{x}_T$ is a $d_T$ dimensional vector.

*4.2. DOM Tree Encoder*

We use GNN as the DOM tree encoder. Firstly, we construct the graph structure using in GNNs. Let $\mathcal{G} = (\mathcal{V}, \mathcal{E})$ denotes the graph feeding to the networks. Obviously, we have $\mathcal{V} = \mathcal{N}$. According to the definition of $\mathcal{R}$, it contains two types of relationships, a parent-child relationship and a sibling relationship. Notice that the relationships in $\mathcal{R}$ are directionless, but GNN's message passing requires the direction of edges. We only use the parent-child relationship here to construct directed edges $v_p \to v_c$ and $v_c \to v_p$, where $v_p$ denotes the parent node and $v_c$ denotes the child node.

Secondly, we build the representation of the node $v \in \mathcal{V}$. XPath is a path string that uniquely identifies an HTML DOM node, and can be used to easily locate a node in the document. We designed our embedding layer by referring to the implementation of the XPath embedding layer in MarkupLM[8]. Given a node $v$, we add up the tag unit embedding and subscript unit embedding to obtain the embedding $ue_j$ of the $j$-th unit. MarkupLM set max depth $L_{xp} = 50$ as default. However, after looking at the length of the XPath of DOM nodes in the dataset we use, we found that the actual length of the sequence obtained by splitting the XPath is much less than 50. Therefore, we take $L_{xp} = 15$. Finally, we concatenate all the unit embeddings to get the representation $h_0$ of XPath of node $v$, i.e., $h_0 = \|_{k=0}^{L} ue_k$. The final embedding of XPath $h$ is obtained as $h = \text{Dropout}\left(\sigma(\text{LayerNorm}(h_0))\right)$, where $\text{LayerNorm}(\cdot)$ stands for the Layer Normalization operation.

For all the nodes in set $\mathcal{V}$, we use GNNs to update the representation of each node. The updating process of the $l$-th GNN layer for each node $v \in \mathcal{V}$ can be described as $m_v^{(l)} = \text{AGGREGATE}(\{h_u^{(l-1)}: u \in N(v)\}$, $h_v^{(l)} = \text{UPDATE}(h_v^{(l-1)}, m_v^{(l-1)})$, where $m_v^{(l)}$ and $h_v^{(l)}$ denote the message vector and the representation of the $l$-th layer, respectively. $N(v)$ represents all the neighbor of node $v$ in graph $\mathcal{G}$. Function $\text{AGGREGATE}(\cdot)$ and $\text{UPDATE}(\cdot)$ represent the aggregation function and the update function used in GNN, respectively.

Hence, we can construct the node feature matrix $\text{H}^{(L_g)}$ by the nodes' representation after $L_g$ GNN layers updating. $\text{H}^{(L_g)} \in \mathbf{R}^{|\mathcal{V}| \times d_G}$, where $d_G$ denotes the dimension of each node's representation. Eventually, we use a readout function on the graph since what we want actually is a representation of the whole graph. The readout function is usually implemented by some pooling methods. Let $\mathcal{P}$ be a readout function, which stands for a pooling strategy. Then we can obtain the whole graph representation $\mathbf{x}_G$ as $\mathbf{x}_G = \mathcal{P}\left(\text{H}^{(L_g)}\right)$, where $\mathbf{x}_G$ is a $d_G$ dimensional vector. We use a pooler to increase the expressiveness of the model and prevent overfitting, which is illustrated as $\mathbf{x}_G = \text{pooler}(\mathbf{x}_G) = \sigma\left(\text{BatchNorm}(\mathcal{F}(\mathbf{x}_G))\right)$, where $\mathcal{F}$ denote a $\mathbf{R}^{d_G} \to \mathbf{R}^{d_G}$ linear layer and $\text{BatchNorm}(\cdot)$ stands for the Batch Normalization operation.

*4.3. Classifier*

In Sections 4.1 and 4.2, we have got the representation of text $\mathbf{x}_T$ and graph $\mathbf{x}_G$, respectively. Since $\mathbf{x}_T$ and $\mathbf{x}_G$ come from different models, to unify them to the same measure, we do normalization on $\mathbf{x}_T$ and $\mathbf{x}_G$, i.e., $\mathbf{x}_T' = \frac{\mathbf{x}_T}{\|\mathbf{x}_T\|_2}$, $\mathbf{x}_G' = \frac{\mathbf{x}_G}{\|\mathbf{x}_G\|_2}$

In order to get the representation $\mathbf{x}_H$ for the whole HTML document, we simply concatenate the two representations as $\mathbf{x}_H = \mathbf{x}_T' \mathbf{x}_G'$, where the dimension $d_H$ of $\mathbf{x}_H$ equals to $d_T + d_G$.

We fed the representation into an MLP for multi-class classification, as illustrated below:

$$o = \text{MLP}(\mathbf{x}_G), \quad o \in \mathbf{R}^{|C|}$$

where $|C|$ denotes the number of elements in the categories set $C$, i.e., the number of pre-defined labels.

Eventually, we apply the $softmax(\cdot)$ function to normalize $o$ and select the maximum probability as the prediction $\hat{y}$. We use the cross-entropy function as the loss function to do optimization.

**5. Experiments**

*5.1. Datasets*
We conduct the experiments on three datasets. KI-04 is a dataset under genre classification. SWDE[9] is a dataset commonly used to test the performance of information extraction(IE). Since SWDE stores web pages under separate categories by domain, we try to use this dataset here for web page classification.
To test the performance of this model on real-world problems, we constructed our own dataset, AHS. It is the abbreviation of *Academic Homepages of Scholars*. The practical problem here is how to automatically perform information extraction on the homepages of scholars from different schools for influence evaluation later. We first crawled teachers' academic homepages of 22 universities, with the purpose of addressing the training data needed for an automatic web page information extraction model. Each university contains four colleges, and each college contains at least about 15 scholars' (or teachers') personal homepages. Based on the above data, we built the 1.0 version of AHS.
After that, we found that in the pre-stage of the whole collection system, we need to determine whether the web page fed to the system is a member's academic homepage indeed. Then this problem evolves into a binary classification task for whether it is an academic homepage or not. Since the AHS 1.0 only contains the personal homepages of scholars, we incrementally crawled another part of the web pages as negative samples due to the above issues. We crawled the Nbaplayer homepage data and movie information homepage data from the web respectively. Based on the above work, we built AHS 2.0.
The information of the datasets on which our experiments depend is shown in Table 1, Cat. is the short of categories.

| Dataset | #Cat. | #Pages | Language |
|---|---|---|---|
| KI-04 | 8 | 1239 | English |
| SWDE | 9 | 124,291 | English |
| AHS 1.0 | 22 | 8937 | Chinese, English |
| AHS 2.0 | 2 | 11244 | Chinese, English |

Table 1: Information of datasets

| Dataset | Acc. | R. | P. | F1 |
|---|---|---|---|---|
| KI-04 | 1.000 | 1.000 | 1.000 | 1.000 |
| SWDE | 0.902 | 0.913 | 0.904 | 0.897 |
| AHS 1.0 | 0.992 | 0.994 | 0.993 | 0.992 |
| **AHS 2.0** | **1.000** | **0.998** | **0.999** | **0.999** |

Table 2: Model performance of PLM-GNN

### 5.2. Implementation Details

We implement our PLM text encoder based on pre-trained models provided in Huggingface Transformers[10]. We use RoBERTa-large as an English text encoder to encode KI-04 and SWDE datasets. At the same time, we use hfl/chinese-RoBERTa-wwm-ext-large as a cross-lingual text encoder to encode the AHS dataset. Secondly, we implement the GNNs relying on the DGL framework. We employ $sum(\cdot)$ on each dimension as a readout function here to get the graph-level representation. Finally, we fed the vector to a 2 layer MLP for classification. We use AdamW as an optimizer with learning rate of 3e-4 to train the whole model.

### 5.3. Results

We evaluate the performance of the model by four metrics, which are accuracy, recall, precision, and Macro F1. The results are shown in Table 2. All the metrics on the KI-04 dataset reach 1.000, which is caused by the small volume. Meanwhile, PLM-GNN performed well on both versions of the AHS dataset.

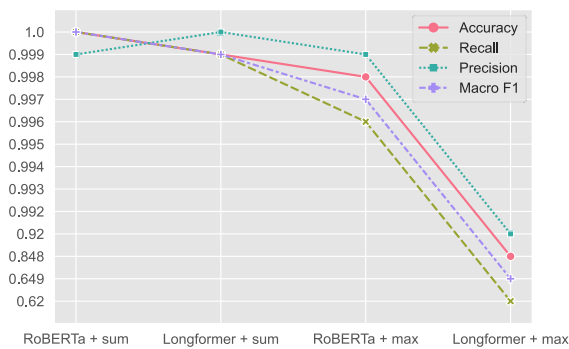

**Figure 2.** Model performance of replacing different encoders.

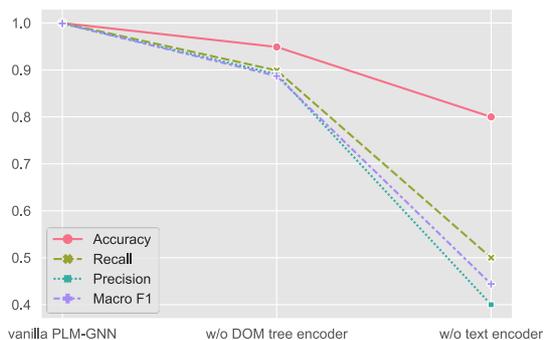

**Figure 3.** Model performance using modules separately.

### 5.4. Ablation Study

To test the effect of different modules in the model, we conducted the following sets of experiments. Since the ultimate goal of this paper is to solve practical problems, the following experiments are oriented to AHS 2.0.

First, we replace the text encoder and readout function of GNN respectively. Some models are proposed to solve the problem of BERT input length limitation, such as $L = 4096$ for Longformer[11].

We replaced RoBERTa with Longformer and the sum(·) readout function with max(·). The results are shown in Figure 2.

It can be seen that replacing RoBERTa with Longformer leads to some performance degradation. We believe this may be due to two reasons. One is that the complexity leads to overfitting of the model, and the other is that it is not necessary to input all the text in the web page to be able to achieve good results. In the DOM tree encoder, the impact of replacing the readout function is relatively small, but there is a slight performance degradation after changing.

Furthermore, we explored the role of these two modules in the overall model by using the text encoder and the DOM tree encoder separately for classification, as shown in Figure 3. According to the results, we can see that the text of the web page is the key feature to distinguish the web page, but the graph structure features also complement the features well.

### 6. Conclusion
In this paper, we propose a simple model for representing and classifying web pages, PLM-GNN. We use a pre-trained language model to encode the text in web pages, and a graph neural network to model the structural information of DOM trees. The model does not require manual feature construction for web pages, but can automatically learn to obtain a representation. With this model, we solve the problem of classifying web pages within the process of building an automatic academic information collection system.